\documentclass[runningheads]{llncs}

\usepackage[year=2026,ID=8851]{eccv} 

\usepackage{datenumber}
\usepackage{calc}
\usepackage[mmddyyyy]{datetime}
\newcounter{datetoday}
\newcounter{diffyears}
\newcounter{diffmonths}
\newcounter{diffdays}
\newcommand{\difftoday}[3]{%
      \setmydatenumber{datetoday}{\the\year}{\the\month}{\the\day}%
      \setmydatenumber{diffdays}{#1}{#2}{#3}%
      \addtocounter{diffdays}{-\thedatetoday}%
      \ifnum\value{diffdays}>0
        \def\diffbefore{}%
        \def\diffafter{left}%
      \else
        \def\diffbefore{}%
        \def\diffafter{ago}%
        \setcounter{diffdays}{-\value{diffdays}}%
      \fi
      \setcounter{diffyears}{\value{diffdays}/365}%
      \setcounter{diffdays}{\value{diffdays}-365*\value{diffyears}}%
      \setcounter{diffmonths}{\value{diffdays}/30}%
      \setcounter{diffdays}{\value{diffdays}-30*\value{diffmonths}}%
      \diffbefore
      \ifnum\value{diffyears}=0
      \else
        \ifnum\value{diffyears}>1
            \thediffyears\space years,
        \else
            \thediffyears\space year,
        \fi
      \fi
      \ifnum\value{diffmonths}=0
      \else
        \ifnum\value{diffmonths}>1
            \thediffmonths\space months
        \else
            \thediffmonths\space month
        \fi
      \fi
      \ifnum\value{diffdays}=0
      \else
        \ifnum\value{diffdays}>1
            \thediffdays\space days
        \else
            \thediffdays\space day
        \fi
      \fi
      \diffafter
}

\newcommand{\NEW}[1]{{{{#1}}}} 
\usepackage{amsbsy}
\usepackage[mathscr]{euscript}
\usepackage{amsmath}
\usepackage{stackrel}
\usepackage{nicefrac}
\usepackage{kotex}

\usepackage{algpseudocode}
\usepackage[ruled]{algorithm2e} 

\SetAlFnt{\small}
\SetAlCapFnt{\small}
\SetAlCapNameFnt{\small}
\SetAlCapHSkip{0pt}
\IncMargin{-\parindent}

\usepackage{dblfloatfix}
\usepackage{float}

\usepackage{subcaption}

\usepackage{booktabs}
\usepackage{multirow}
\usepackage[table]{xcolor}
\usepackage{makecell}
\usepackage{array}
\usepackage{tabularx}

\definecolor{good}{RGB}{198,239,206}   
\definecolor{mid}{RGB}{255,242,204}    
\definecolor{bad}{RGB}{255,199,206}    

\newcolumntype{L}[1]{>{\raggedright\arraybackslash}p{#1}}

\usepackage{graphicx}
\usepackage{amsmath,amssymb}

\title{Depth-guided Multi-view Exposure Bracketing for HDR Robot Vision}

\author{
Jinnyeong Kim ~ ~ Juhyung Choi ~ ~ Woohyeok Kim ~ ~ Sunghyun Cho ~ ~ Seung-Hwan Baek 
}

\institute{
POSTECH
}
\begin{document}
\maketitle

\begin{abstract}
Achieving reliable single-shot high dynamic range (HDR) imaging under extreme illumination conditions remains a long-standing challenge, 
yet no comprehensive benchmark exist for evaluating HDR perception in multi-sensor robotic systems. 
To fill this gap, we introduce a large-scale dataset collected via a custom robotic vision platform and an iPhone 13 Pro: 121 real-world scenes spanning modest and ultra-high dynamic range conditions, alongside 20 synthetic video sequences from the CARLA simulator.
As a reference pipeline for this dataset, we propose Depth-guided Multi-view Exposure Bracketing (DMEB), a single-shot HDR method that distributes drastically different exposures across multi-view low-bit-depth cameras and fuses them via depth-guided confidence-aware fusion. 
Evaluations on our dataset show that DMEB establishes a strong reference point and highlight the promise of this sensor configuration for robust HDR perception in diverse multi-camera and depth sensor system.
\end{abstract}
\newcommand{\new}[1]{\textcolor{blue}{#1}}

\section{Introduction}
\label{sec:intro}

Modern robotic and mobile platforms increasingly \NEW{carry multiple cameras and depth sensors. To operate reliably under extreme lighting, these platforms require high dynamic range (HDR) imaging capabilities.
}
Temporal exposure bracketing is the representative method, reconstructing an HDR image by fusing multiple exposures captured sequentially by a single camera~\cite{debevec2023hdr}. However, for dynamic scenes, sequential capture introduces motion artifacts~\cite{tursun2015hdrghostreview}. Although these artifacts can be partially compensated for by estimating optical flow between differently exposed images~\cite{sun2010stereodisphdr,zhang2025hdrstereovideo,choi2025dual} , robust flow estimation unfortunately requires the use of similar exposure levels in the exposure bracketing, which significantly limits the achievable dynamic range expansion. Consequently, specialized, high-end cameras with high-bit-depth sensors are often employed albeit substantially increasing system costs~\cite{LUCID_2023_TritonHDR_AltaView,Sony_IMX490_2019,eConSystems_STURDeCAM31_NVIDIAList_2025}.

\NEW{
The multi-sensor configuration~\cite{menze2015object,geiger2012we} already present on many platforms suggests a different direction.} 
\NEW{
In many perception tasks, depth is fused with RGB images to provide geometric cues across views, such as 3D object detection~\cite{mao20233d} and LiDAR-assisted 3D reconstruction~\cite{cui2024letsgo}.
For HDR reconstruction, such geometry can be especially useful because photometric alignment becomes unreliable as exposure differences across views grow.}

\begin{figure}[t]
  
    \includegraphics[width=\linewidth]{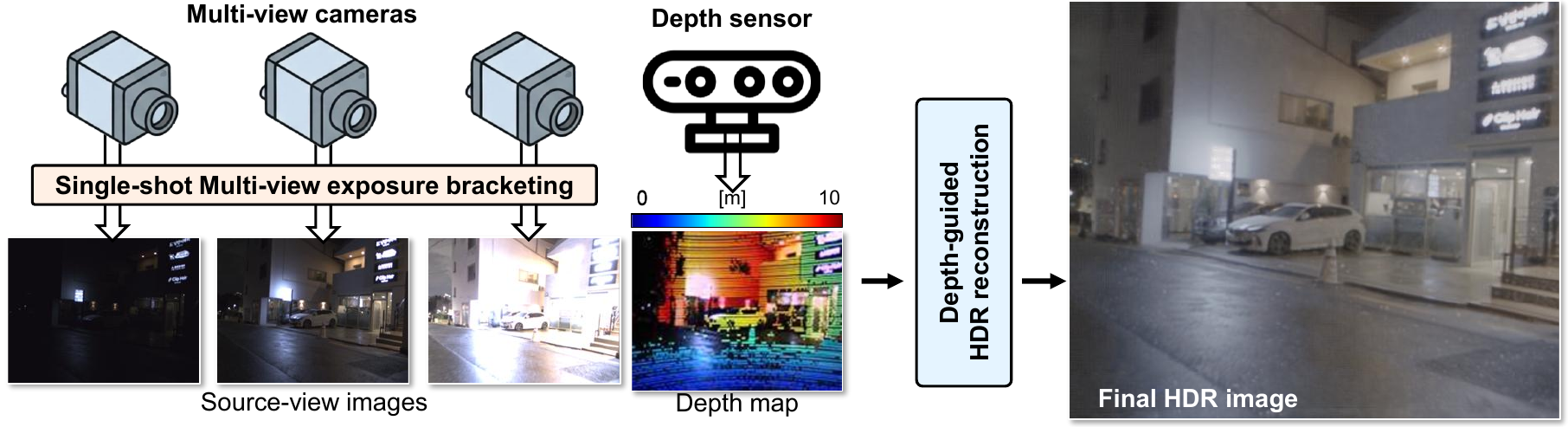}
\captionof{figure}{
\textbf{Overview.}
To effectively expand the achievable dynamic range, DMEB distributes drastically different exposures across multi-view cameras, facilitating robust, real-time HDR reconstruction through depth-guided fusion.
}
    \label{fig:overview}
\end{figure}
\NEW{To our knowledge, no public dataset provides calibrated multi-view images captured simultaneously with drastically different exposures and paired with aligned depth, making it difficult to evaluate spatially distributed exposure bracketing as an alternative to temporal bracketing and specialized HDR sensors.}
We address this gap by constructing a dataset for multi-view varying-exposure HDR reconstruction with aligned depth. The dataset spans three complementary sources: a robotic vision dataset for controlled real-world evaluation across camera counts and depth modalities, an iPhone dataset for compact consumer devices, and a synthetic dataset with rendered HDR ground truth for extreme-DR stress testing. Together, these sources provide synchronized multi-view images with complementary exposures and depth across real, consumer, and simulated settings, establishing a testbed that was previously unavailable.

\NEW{As a baseline method for this dataset, we introduce Depth-guided Multi-view Exposure Bracketing (DMEB).}
DMEB reconstructs an HDR image in a single shot by distributing exposure settings across synchronized cameras rather than across time. It uses aligned depth, such as LiDAR or active stereo, to warp differently exposed images into a reference view and performs confidence-aware fusion. This geometry-guided formulation allows substantially larger exposure differences than photometric matching alone, thereby expanding the recoverable dynamic range within a single capture.


\NEW{
Evaluations on the dataset show that calibrated multi-view cameras with complementary exposures, together with depth, provide a practical alternative to temporal bracketing and specialized HDR sensors for single-shot HDR perception. 
The datasets and code are publicly available at \url{https://divisonofficer.github.io/dmeb}.

}

Our technical contributions are summarized as follows:
\begin{itemize}
\item \NEW{We introduce a multi-view varying-exposure HDR dataset collected across three platforms---a robotic vision rig, a consumer smartphone, and a synthetic renderer---with synchronized images captured under drastically different exposures and paired with calibrated depth measurements, enabling evaluation across camera counts and depth modalities.}

\item \NEW{We provide Depth-guided Multi-view Exposure Bracketing (DMEB) as a baseline method for this dataset, replacing temporal bracketing with depth-guided multi-view fusion for single-shot HDR reconstruction under large exposure gaps.}

\end{itemize}

\section{Related Work}
\label{sec:related}


\paragraph{Exposure Bracketing.}
Exposure bracketing captures multiple LDR images with varying exposures from a single camera and fuses them into an HDR image~\cite{debevec2023hdr,debevec2000acquiring,granados2010optimal,MitsunagaNayar2000,Robertson2003PAMI,Tsin2001CVPR}. While this method performs well for stationary scenes and cameras, it struggles with dynamic scenes, as sequentially captured frames exhibit motion-induced misalignment. To mitigate this issue, prior works align frames via correspondence matching using patches~\cite{sen2012robusthdrpatch}, flow~\cite{zimmer2011freehandhdrflow,kalantari2013patchflow,kalantari2019deepflow,catley2022flexhdrflow,kong2024safnet,xu2024hdrflow}, or, more recently, attention~\cite{yan2019attentionhdr,pu2020robusthdrattention,liu2022hdrtransformer,song2022selectivetransformer,yan2023unifiedhdrtransformer}. However, correspondence estimation between images with drastically different exposures is fundamentally unreliable. Consequently, these methods resort to using similar exposures between frames, which severely limits the achievable dynamic range expansion. In contrast, DMEB successfully leverages vastly different exposures across multi-view cameras to achieve substantial dynamic range expansion.

\paragraph{Multi-camera HDR Imaging.}
Previous research considered distributing different exposures across multiple cameras~\cite{sun2010stereodisphdr,popovic2016multi,zhang2025hdrstereovideo,li2023dualhdr,choi2025dual}.
However, because differently-exposed images need to be aligned, they restrict the exposure differences across cameras to maintain photometric multi-view consistency, which in turn limits dynamic range expansion.
DMEB overcomes this limitation by leveraging depth obtained from an external sensor, allowing for using drastically-different exposures for effective dynamic range expansion.

\paragraph{Specialized Cameras for HDR Imaging.}
Specialized sensors and optics directly enhance per-frame dynamic range. 
Pixel- or readout-level approaches, including multi-exposure Bayer patterns~\cite{LUCID_2023_TritonHDR_AltaView,Sony_IMX490_2019,eConSystems_STURDeCAM31_NVIDIAList_2025}, variable quantization schemes~\cite{liu20204multiquant}, and multi-gain readouts~\cite{lee2025adaptivemultigain,hajisharif2015adaptive,takayanagi2018over,park2022world,cho2011alternating}, integrate single-shot HDR capability into sensor hardware. However, they suffer from amplified noise, limited range expansion, and high system cost.
Beam-splitter-based multi-camera systems sacrifice half of the incoming light and difficult to scale beyond two cameras~\cite{kronander2014unifiedbeam,tocci2011versatilehdrbeamsplit}.
PSF-engineered optics for HDR imaging encode intensity variations across the PSF, enabling single-shot HDR reconstruction~\cite{metzler2020deep,brookshire2024metahdr,sun2020learning}.
However, they sacrifice image spatial resolution.
In contrast, DMEB offers robust single-shot HDR utilizing a sensor suite that is becoming ubiquitous in robots and mobile devices: standard multi-view, low-bit-depth cameras paired with a depth sensor.

\paragraph{Datasets for HDR Imaging.}
While the widely used HDR benchmarks rely on exposure bracketing, they are primarily designed for single-camera setups~\cite{Kalantari2017DeepHDR,Liu_2023_CVPR,Zou_2023_ICCV,Chen_2021_ICCV,Shu_2024_CVPR}. Although a public HDR dataset for stereo cameras exists~\cite{choi2025dual}, it does not contain images with varying exposures between cameras at a time. Our datasets are constructed to enable testing of DMEB that require varying-exposure multi-view cameras. Moreover, it supports testing DMEB with varying number of cameras and diverse depth-sensing modalities.

\setlength{\belowcaptionskip}{-3mm}

\section{Depth-guided Multi-view Exposure Bracketing}
{
Given synchronized low-bit-depth images captured with complementary exposures and an aligned depth measurement, DMEB reconstructs an HDR image in a selected reference-camera view. The pipeline consists of three steps: multi-view exposure control, depth- and confidence-guided fusion, and HDR refinement.
}

\subsection{Multi-view Exposure Control}
\begin{figure}[t]
    \centering
    \includegraphics[width=\linewidth]{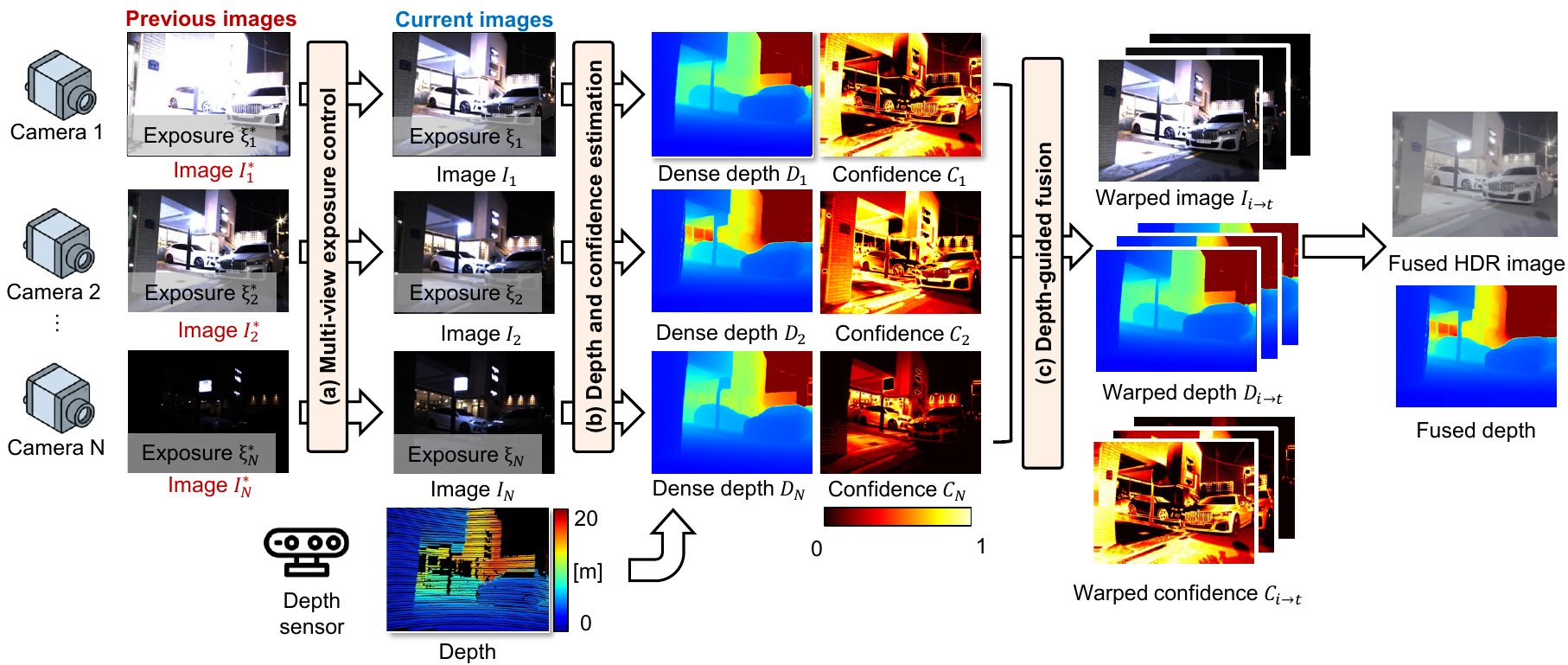}
    \caption{\textbf{Exposure control and HDR reconstruction.}
(a) By analyzing previous-frame images, we control exposures of multi-view cameras. (b) We densify the depth measurements and compute confidence for fusion. (c) Guided by the dense depth, we warp the source images, depths, and confidences to the reference view, performing a confidence-aware fusion to reconstruct the final HDR image and fused depth map. }
    \label{fig:dmeb}
\end{figure}

Figure \ref{fig:dmeb}(a) illustrates our multi-view exposure control strategy. Given the previous-frame multi-view images $\{I_i^*\}_{i=1}^N$ and their corresponding exposure settings $\{\xi_i^*\}_{i=1}^N$, we determine the current-frame exposure settings $\{\xi_i\}_{i=1}^N$ across the $N$ multi-view cameras.
To adapt to illumination changes, we update the exposure $\xi_i$ for the $i$-th camera to achieve a target mean intensity $m_i$:
\begin{equation}
    \xi_i \leftarrow \operatorname{clip}\!\left(\xi_i \frac{m_i}{m_i^*}, \xi_{\min}, \xi_{\max}\right),
\end{equation}
where $m_i^*$ is the mean intensity of the previous-frame image from the $i$-th camera, and $\operatorname{clip}(\cdot)$ restricts the exposure within the sensor-defined hardware limits, $\xi_{\min}$ and $\xi_{\max}$.
The target mean intensities are geometrically spaced between a low and high target brightness, with $m_1=0.05$ and $m_N=0.8$:
\begin{equation}
    \log m_i = \log m_1 + \frac{i-1}{N-1}\big(\log m_N - \log m_1\big).
\end{equation}

Given the computed exposure settings $\{\xi_i\}_{i=1}^N$, we capture the corresponding LDR images $\{I_i\}_{i=1}^N$ from the multi-view cameras and simultaneously acquire depth $D$ from the depth sensor. 
The exposure $\xi_i$ is set to the camera by converting the exposure value into exposure time $\tau_i$ and gain $g_i$ as 
\begin{equation}
    \tau_i = \min(\xi_i, \tau_{\max}), \quad g_i = \max\big(1, \min(g_{\max}, \xi_i / \tau_i)\big),
\end{equation}
where $\tau_{\max}$ denotes the maximum stable exposure time supported by the system (approximately 50 ms for 8 FPS operation), and $g_{\max}$ is the sensor-imposed upper gain limit.

\subsection{Depth-guided HDR Reconstruction}
Using the depth maps and images, we reconstruct an HDR image $H_t$ for a target view $t$, which we set to be $t=N/2$ in our experiments.

\paragraph{Dense Depth.}
We first warp the raw depth $D$ to each camera, producing a per-camera depth map $D_i$. Next, we estimate a dense depth map $\hat{D}_i$ utilizing a pretrained monocular depth prior~\cite{yang2024depthanything}, while preserving the metric scale guided by the sparse depth $D_i$ via scale estimation, inspired by previous work~\cite{fan2025region}. 
Figure~\ref{fig:dmeb}(b) shows the dense depth.

\paragraph{Confidence Estimation.}
We compute a per-pixel confidence map $C_i$ to downweight unreliable measurements (e.g., saturation, under-exposure noise, and occlusions) during the HDR reconstruction process. We estimate this confidence as:
\begin{equation}
    C_i = f_{\text{trapezoid}}(I_i) \cdot f_{\text{CNN}}(I_i, \hat{D}_i),
\end{equation}
where $f_{\text{trapezoid}}$ is an intensity-based trapezoid function~\cite{debevec2023hdr} and $f_{\text{CNN}}(I_i, \hat{D}_i)$ is a convolutional neural network detailed in the Supplementary Material.
Figure~\ref{fig:dmeb}(b) shows the estimated confidence maps.

\paragraph{Depth-guided Fusion.}
We use the dense depth $\hat{D}*t$ to warp the per-view depth map, confidence map, and image to a fixed reference view $t$. This results in aligned sets of images ${I*{i\rightarrow t}}*{i=1}^N$, depths ${\hat{D}*{i\rightarrow t}}*{i=1}^N$, and confidences ${C*{i\rightarrow t}}_{i=1}^N$, as shown in Fig.~\ref{fig:dmeb}(c).
We then merge the warped depth maps and images into a single fused depth $\dot{D}_t$ and an HDR image $H_t$ as

\begin{equation}
\dot{D}_t = \frac{\sum_i C_{i\rightarrow t}\, \hat{D}_{i\rightarrow t}}{\sum_i C_{i\rightarrow t} + \epsilon}, \quad H_t = \frac{\sum_i (C_{i\rightarrow t}\, V_{i\rightarrow t})\, \frac{I_{i\rightarrow t}}{ \tau_i g_i}}{\sum_i (C_{i\rightarrow t}\, V_{i\rightarrow t}) + \epsilon},
\end{equation}
where $\epsilon$ is a small constant and $V_{i\rightarrow t}$ is a soft visibility mask computed via depth consistency:
\begin{equation}
V_{i\rightarrow t} = \sigma\!\left(
        \frac{(\dot{D}_t + \delta) - \hat{D}_{i\rightarrow t}}
             {\tau_{\mathrm{rel}} \cdot (\dot{D}_t + \varepsilon)}
    \right) + w_\text{floor},
\end{equation}
Here, $\delta$ is a depth margin (set to $1\,\mathrm{m}$ in all experiments) to absorb cross-view misalignment. The parameter $\tau_{\mathrm{rel}}$ controls the tolerance to relative depth variations, $\varepsilon$ avoids division by zero, and $\sigma(\cdot)$ denotes the logistic function. A small constant $w_\text{floor}=0.05$ ensures that no pixels are entirely excluded from the fusion. \NEW{The reconstructed HDR image is defined in the reference-camera view point. During warping, a source view contributes only where its projection is valid in the source image and geometrically consistent with the reference view. Invalid or unreliable observations are therefore assigned low weight through the confidence and visibility terms before refinement.}
Figure~\ref{fig:dmeb}(c) \NEW{illustrates this fusion step, which produces the fused HDR image $H_t$ and depth $\dot{D}_t$.}
Refer to Sec.~2 of the Supplementary Material for further details.

\paragraph{Refinement.}
To refine the fused HDR image $H_t$, we map it to a tonemapped space using the $\mu$-law tone mapper~\cite{Kalantari2017DeepHDR}. To properly handle the extreme dynamic range expansion achieved by our method, we generate three distinct tonemapped images using $\mu_1 = 10^3$, $\mu_2 = 5\times10^4$, and $\mu_3 = 10^6$, representing different exposure levels. To combine these three tonemapped images, we employ a transformer network. The output HDR image is converted back to the linear intensity domain using the inverse $\mu$-law function, resulting in our final HDR estimate. Refer to the Supplementary Material for details.

\setlength{\belowcaptionskip}{-3mm}

\begin{figure*}[t]
    \includegraphics[width=\textwidth]{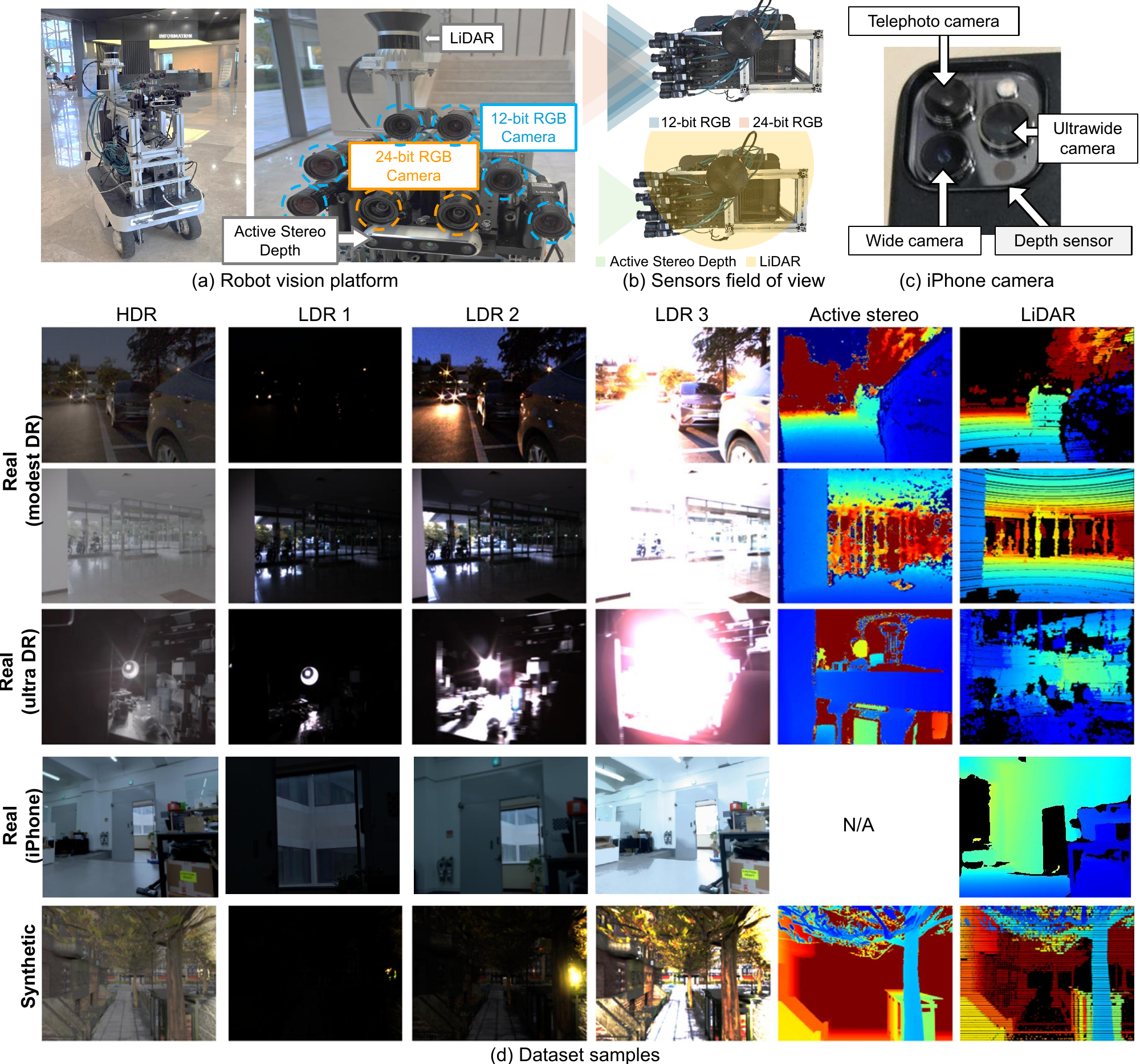}
    \captionof{figure}{
\textbf{Systems and Datasets.} We test DMEB on two real-world systems: (a) a robotic vision platform featuring multiple sensors with (b) shared fields of view, and (c) an iPhone equipped with three cameras and LiDAR. (d) Using these systems and a synthetic renderer, we present datasets consisting of multi-view LDR images with varying exposures, HDR reference images, and depth maps.}
    \label{fig:system_and_dataset}
\end{figure*}


\paragraph{Training.}
Our trainable components consist of the confidence-estimation network and the HDR refinement network. We train these models using our newly proposed datasets, detailed in the following section, applying a 9:1 train/test split. The training process begins with a pretraining phase on 40k synthetic scenes for 50 epochs. In this stage, we optimize based on both the final HDR and depth estimates; the HDR reconstruction is supervised via an $\ell_1$ loss, SSIM, and a gradient loss within the $\mu$-law tone-mapped domain~\cite{Kalantari2017DeepHDR}, whereas the depth reconstruction utilizes an $\ell_1$ loss alongside a SILog loss~\cite{eigen2014depth}. Following this, we fine-tune both networks on 10k real scenes for 10 epochs, relying on the HDR reconstruction losses. All optimizations are carried out using Adam with a learning rate of $10^{-4}$.

\section{Datasets}
\label{sec:inverse_rendering}

\subsection{Robot Vision Dataset}
\paragraph{Setup.}
To evaluate the proposed DMEB framework, we constructed the custom robotic vision platform shown in Figure~\ref{fig:system_and_dataset}(a). 
While DMEB requires a minimum of only two cameras, our platform integrates six 12-bit LDR cameras (Lucid Triton TRI032S-CC) to facilitate testing with varying camera counts, along with two 24-bit HDR cameras (Lucid Triton TRI054S) that serve as pseudo-ground truth references. Furthermore, we equipped the system with an active-stereo depth sensor (Intel RealSense D455) and a LiDAR (Ouster OS1). These provide complementary depth characteristics, allowing us to analyze the impact of the chosen depth modality. All sensors are synchronized, geometrically calibrated, and configured with overlapping fields of view. Figure~\ref{fig:system_and_dataset}(b) visualizes the field-of-view coverage of each camera and depth sensor in our setup. The imaging system is mounted on a Ranger Mini 2.0 four-wheeled mobile base and operated via a laptop. Further system and calibration details are provided in the Supplementary Material.

\paragraph{Dataset.}
Using the robotic vision platform, we collected two types of real-world datasets, which are the main datasets for real-world evaluation. The real (modest DR) dataset contains 31 scenes and 15,000 frames captured while the robot was in motion. The real (ultra DR) dataset comprises 80 static scenes captured without robot motion, providing high-quality ground-truth HDR images for each camera view using temporal exposure bracketing. Figure~\ref{fig:system_and_dataset}(d) shows sample data.
Please refer to the Supplementary Material for detailed statistics of the dataset.

\subsection{Synthetic Dataset}
Beyond serving as training data, our synthetic dataset functions as an extreme-ground-truth benchmark: it provides rendered HDR reference images, metric depth, and sensor-like depth inputs under illumination ranges that are difficult to capture reliably with real hardware.

We created this dataset using the CARLA simulator~\cite{Dosovitskiy17carla}, rendering multi-view HDR images and depth maps across six simulation environments. These scenes span diverse lighting conditions (day, dusk, and night) with randomized weather, traffic, and spatial layouts, resulting in 20 video sequences of 15{,}000 frames in total.

Figure~\ref{fig:system_and_dataset}(d) shows sample synthetic data.
We store the rendered HDR frames in the same EXR format and camera configuration as our real dataset, using an 8-camera rig with matched intrinsics/extrinsics.
The simulator provides ground-truth metric depth for every view, and we additionally render sensor-like inputs including sparse LiDAR depth and active-stereo depth maps to mirror our real capture pipeline.

\NEW{This synthetic dataset allows us to test HDR reconstruction under controlled exposure spacing, depth noise, and scene motion. In particular, the rendered HDR ground truth exceed $150$\,dB effective dynamic range, making it a testbed for extreme-DR evaluation.}

\subsection{iPhone Dataset}
To demonstrate that DMEB can be applied to everyday consumer devices, we also acquired a dataset using an iPhone 13 Pro, which features three LDR cameras (wide, ultrawide, and telephoto) and a LiDAR sensor. Using this setup, we captured a small-scale dataset containing 10 scenes. 
Figure~\ref{fig:system_and_dataset}(c) illustrates the camera specification of iPhone device that is used for our iPhone dataset.
Figure~\ref{fig:system_and_dataset}(d) shows sample data.
For each scene, we capture exposure-bracketed bursts on all three cameras by sweeping the shutter time over six levels from 0.01\,ms to 100\,ms, producing LDR inputs at multiple exposure settings.
We construct a reference HDR image by merging each burst~\cite{debevec2023hdr}, which serves as ground-truth HDR supervision.
Each sample additionally includes the iPhone LiDAR measurement, providing a low-resolution depth map aligned to the camera views.

\section{Results}
\label{sec:results}

\begin{table*}[!t]
\setlength{\belowcaptionskip}{-3mm}
\centering
  
\small
\resizebox{\textwidth}{!}{
\begin{tabular}{l l c|ccc|ccc}
\toprule
\multirow{2}{*}{\textbf{Bracketing}} & \multirow{2}{*}{\textbf{Method}} & \multirow{2}{*}{{\textbf{Speed (FPS)} $\uparrow$}} &
\multicolumn{3}{c|}{Real (Modest DR)} &
\multicolumn{3}{c}{Real (Ultra DR)} \\
\cmidrule(lr){4-6}\cmidrule(lr){7-9}
 &  &  & PSNR-$\mu$ & SSIM-$\mu$ & HDR-VDP &
                 PSNR-$\mu$ & SSIM-$\mu$ & HDR-VDP \\
\midrule
\multirow{5}{*}{\makecell{Multi-shot\\ single-camera}}
 & HDR Transformer~\cite{liu2022hdrtransformer} & 1.189
    & 31.26 & 0.859 & 9.64
    & N/A & N/A & N/A \\
 & SAFNet~\cite{kong2024safnet} & 53.937
    & 25.04 & 0.633 & 6.32
    & N/A & N/A & N/A \\
 & HDRFlow~\cite{xu2024hdrflow} & 77.519
    & 28.89 & 0.557 & 7.75
    & N/A & N/A & N/A \\
 & AFUNet~\cite{li2025afunet} & 0.795
    & 29.64 & 0.806 & 9.18
    & N/A & N/A & N/A \\
 & DMEB (ours) & 13.073
    & 30.72 & 0.833 & 9.45
    & N/A & N/A & N/A \\
\midrule
\multirow{5}{*}{\makecell{Single-shot\\ multi-camera}}
 & HDR Transformer~\cite{liu2022hdrtransformer} & 1.189
    & 33.61 & 0.872 & 9.64   
    & 38.74 & 0.848 & 9.40 \\	
 & SAFNet~\cite{kong2024safnet} & 53.937
    & 30.96 & 0.644 & 7.72 
    & 27.39 & 0.769 & 9.06 \\	
 & HDRFlow~\cite{xu2024hdrflow} & 77.519
    & 27.09 &	0.541 &	6.29 
    & 27.44 & 0.617 & 8.95 \\	
 & AFUNet~\cite{li2025afunet} & 0.795
    &32.25 &0.819 &	9.18 
    & 34.69 & 0.877 & 9.39 \\	
 & \textbf{DMEB (ours)} & 13.073
    & \textbf{39.40} &	\textbf{0.916} &	\textbf{9.69}  
    & {39.17} & {0.868} & {9.16} \\
\bottomrule
\end{tabular}}

\caption{\textbf{Quantitative comparison.}
We compare multi-shot single-camera and synchronized single-shot
multi-camera configurations using three inputs.
}
\label{tab:comp_quant_main_reduced}
\end{table*}

\begin{figure*}[t]
    \includegraphics[width=\textwidth]{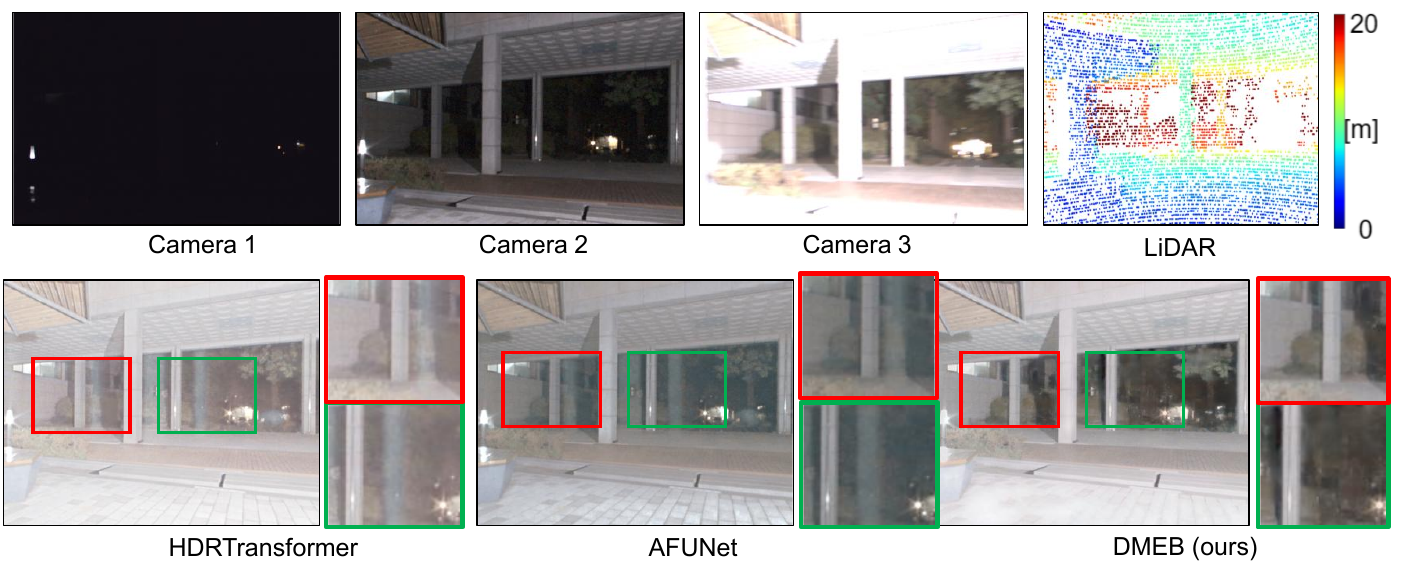}
\captionof{figure}{
\textbf{Comparison with Matching-based HDR.}
Given three images captured with drastically different exposures alongside LiDAR depth, DMEB enables accurate HDR reconstruction. In contrast, HDRTransformer and AFUNet result in reconstruction artifacts.
}
\label{fig:comp_qual}
\end{figure*}

\paragraph{Comparison.}
We compare DMEB against state-of-the-art exposure-bracketing HDR reconstruction networks that take three differently exposed inputs~\cite{liu2022ghost, xu2024hdrflow, kong2024safnet, li2025afunet}. 
All baselines are trained from scratch using the same train/test split, resolution, exposure setting, valid masks, and metric domain.
We assess performance under two configurations:
(1) \textit{multi-shot single-camera} reconstruction using sequential frames synthesized from ground-truth HDR at varying exposures, and
(2) \textit{single-shot multi-camera} reconstruction using synchronized multi-view inputs with different exposures.

Table~\ref{tab:comp_quant_main_reduced} summarizes quantitative results for both configurations on the real datasets using three input images. 
DMEB outperforms previous methods in most evaluation metrics including PSNR, SSIM, and HDR-VDP, while sustaining over 10\,FPS reconstruction speed, enabling reliable HDR imaging for dynamic scenes. 
Additional results for alternative exposure settings and synthetic scenes are included in  Sec.~S.6 of the Supplementary Material.

Figure~\ref{fig:comp_qual} presents a qualitative comparison under the single-shot multi-camera setting with HDRTransformer~\cite{liu2022hdrtransformer}, a baseline that offers strong reconstruction accuracy but operates at only about 1\,FPS. 
All HDR results are shown using a fixed tone-mapping operator.
HDRTransformer produces artifacts due to large exposure gaps across views. 
In contrast, DMEB delivers higher-fidelity reconstruction and substantially broader DR recovery even with only three cameras.

\begin{figure}[!t]
    \centering
    
    \includegraphics[width=\linewidth]{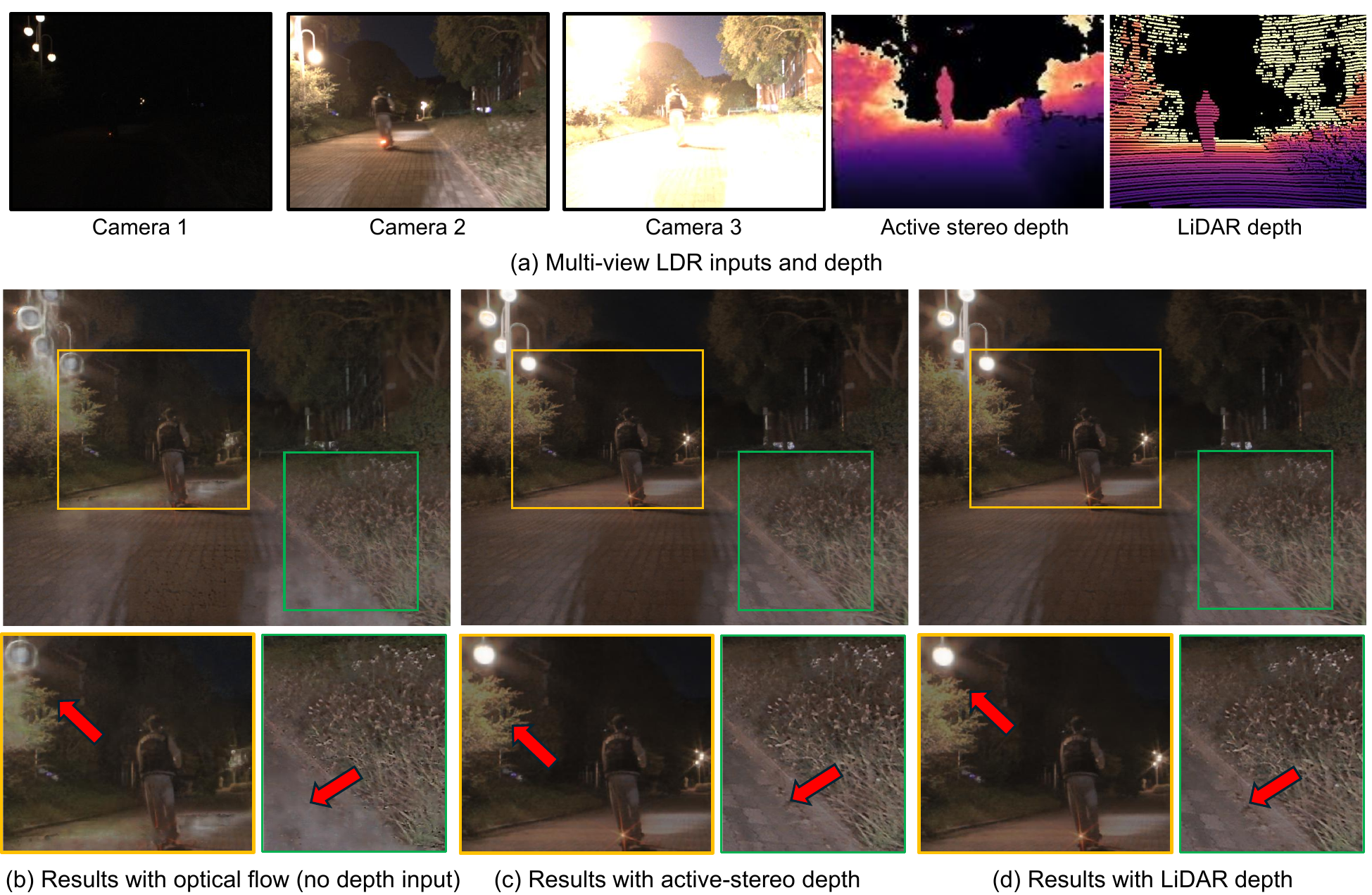}
    \caption{\textbf{Importance of depth and evaluation with different depth modalities.}
DMEB provides robust reconstruction accuracy using either active-stereo or LiDAR depth. However, when depth is not provided and optical flow is used instead for photometric matching, HDR reconstruction fails due to the drastically different exposure gaps between cameras.
}
    \label{fig:depth_modality}
\end{figure}

\begin{table}[t]
\centering
\small
\begin{tabular*}{\linewidth}{@{\extracolsep{\fill}}lcccc@{}}
\toprule
\textbf{Metric} & Flow & Mono & LiDAR-SV & \textbf{LiDAR-MV (ours)} \\
\midrule
PSNR-$\mu$ & 35.84 & 37.09 & 37.90 & \textbf{39.40} \\
\bottomrule
\end{tabular*}
\caption{\textbf{Geometry guidance ablation on the real (modest DR) dataset.}
We compare optical flow without depth, monocular depth, single-view LiDAR-anchored depth, and our full multi-view LiDAR-guided fusion.
SV and MV denote single-view and multi-view fusion, respectively.}
\label{tab:depth_ablation}
\end{table}

\begin{figure}[!t]
    \centering
  
    \includegraphics[width=\linewidth]{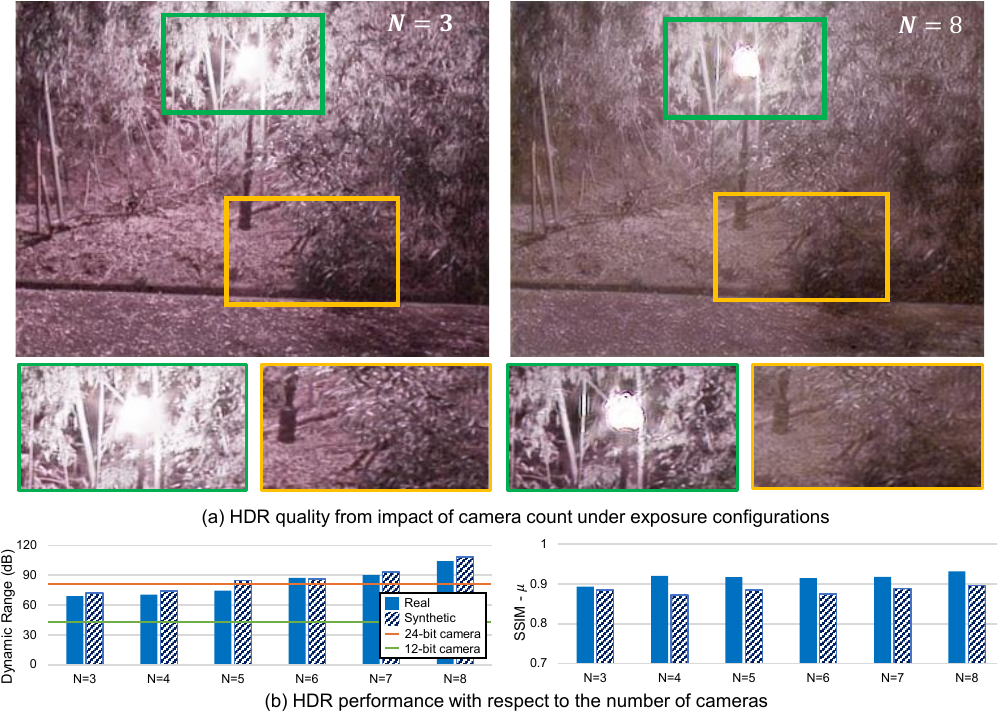}
\caption{\textbf{Impact of camera count.} (a) Qualitative comparison of HDR reconstruction using $N=3$ and $N=8$ cameras. With three cameras, the bright light source remains saturated, whereas expanding the system to eight cameras successfully resolves the underlying details of the light source. (b) Quantitative evaluation demonstrates that the achievable dynamic range expands with the number of cameras, while structural consistency (SSIM) remains stable.}
\label{fig:camera_scalability}
\end{figure}

\begin{figure}[t]
    \centering
    \includegraphics[width=\linewidth]{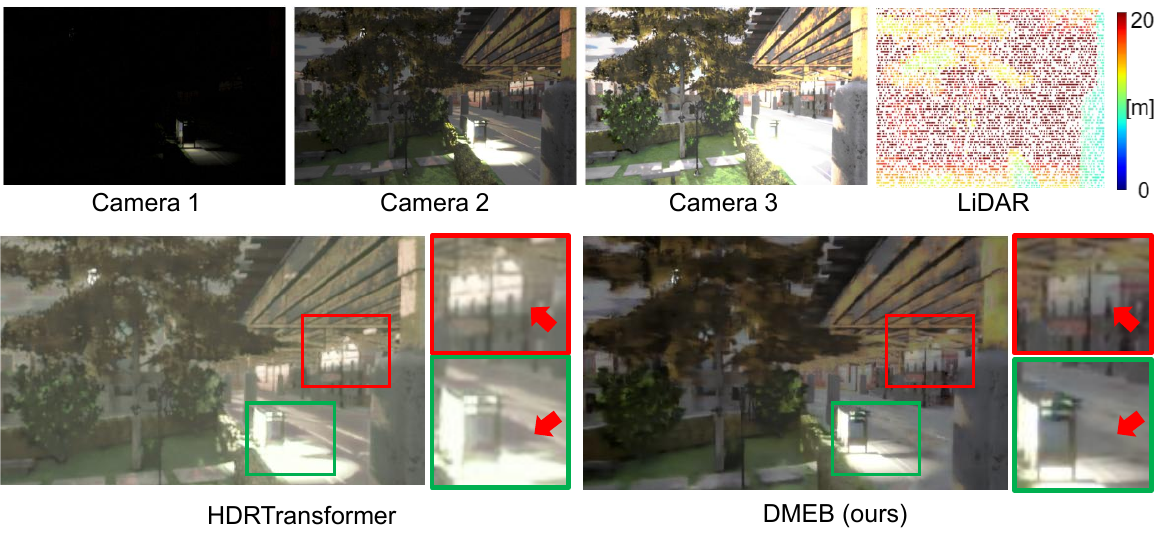}
    \caption{\textbf{Evaluation on the synthetic dataset.}
    { Qualitative comparison between HDRTransformer~\cite{liu2022hdrtransformer} and DMEB (ours) under the single-shot three-view setting.
Red and green boxes indicate cropped regions; zoom-ins are shown on the right, and arrows highlight DMEB outperforms HDRTransformer.
}
    }
    \label{fig:synthetic_result}
\end{figure}

\begin{figure}[t]
    \centering
    \includegraphics[width=\linewidth]{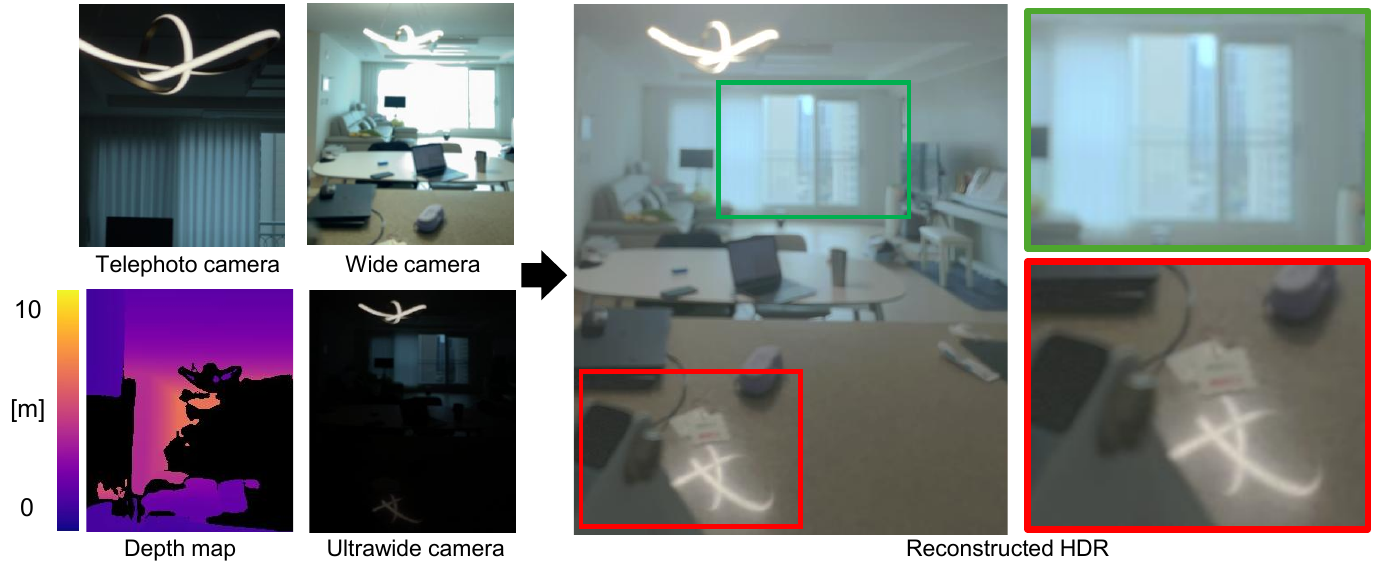}
    \caption{\textbf{Single-shot HDR imaging using an iPhone.}
    By leveraging the built-in sensors of an iPhone~13 Pro (telephoto, wide, and ultrawide cameras, plus LiDAR), DMEB improves reconstruction quality of HDR image. Illustrated here from the ultrawide viewpoint for HDR reconstruction, our method demonstrates potential for high-quality single-shot HDR imaging on consumer electronics.
    }
    \label{fig:process_on_mobile_phone}
\end{figure}

\paragraph{Single-shot Multi-camera Inputs vs. Multi-shot Single-camera Inputs.}
\NEW{Table~\ref{tab:comp_quant_main_reduced} allows us to compare the same HDR reconstruction methods under two input settings: multi-shot single-camera and synchronized single-shot multi-camera. 
For HDRTransformer, changing only the input setting improves PSNR-$\mu$ from 31.26 to 33.61,dB; for AFUNet, it improves from 29.64 to 32.25,dB. This indicates that single-shot multi-camera setup is beneficial beyond DMEB itself. DMEB further leverages this setting with depth-guided fusion, reaching 39.40,dB in the single-shot multi-camera setting.
}
\paragraph{Impact of Depth Guidance.}
Fig.~\ref{fig:depth_modality} shows that both active-stereo and LiDAR depth provide reliable geometric guidance, whereas optical flow struggles to align multi-view images with large exposure differences. Tab.~\ref{tab:depth_ablation} further reports a geometry-source ablation on the real (modest DR) dataset, showing a consistent PSNR-$\mu$ improvement as the depth source moves from photometric correspondence (optical flow) to metric depth, peaking with full multi-view LiDAR-guided fusion. These results indicate that metric depth, rather than photometric correspondence alone, is essential for robust multi-view exposure fusion.

\paragraph{Number of Cameras.}
We investigate the impact of scaling the number of cameras on the overall HDR reconstruction performance of our proposed DMEB framework. To achieve this, we vary the camera count from three to eight by combining images from the six LDR cameras with two LDR-converted images derived from the HDR cameras (details regarding this conversion are provided in the Supplementary Material). Figure~\ref{fig:camera_scalability}(b) quantitatively demonstrates that the achievable dynamic range expands with the addition of cameras, which will plateau when hardware-imposed shutter limits are reached across both real and synthetic datasets. Concurrently, the structural integrity, measured via SSIM, is maintained. From a qualitative perspective, Figure~\ref{fig:camera_scalability}(a) shows that a three-camera setup ($N=3$) suffers from saturation in intensely illuminated regions. In contrast, scaling the system to eight cameras ($N=8$) effectively recovers the underlying details of the bright light source.




\begin{table}[t]
\centering
\small
\begin{tabular*}{\linewidth}{@{\extracolsep{\fill}}lcccc@{}}
\toprule
\textbf{Method} & FPS$\uparrow$ 
 & PSNR-$\mu$ & SSIM-$\mu$ & HDR-VDP \\
\midrule
HDR Transformer~\cite{liu2022hdrtransformer} & 1.19 & 32.65 & 0.830 & 8.20 \\
  SAFNet~\cite{kong2024safnet}               & 53.94 & 28.21 & 0.687 & 6.25 \\
  HDRFlow~\cite{xu2024hdrflow}               & 77.52 & 26.33 & 0.475 & 7.12 \\
  AFUNet~\cite{li2025afunet}                 & 0.80 & 33.52 & 0.830 & 8.02 \\
  \textbf{DMEB (ours)} & 9.76 & \textbf{34.72} & \textbf{0.885} & \textbf{8.29} \\
\bottomrule
\end{tabular*}
\caption{\textbf{Quantitative comparison on the synthetic benchmark.}
The synthetic ground truth exceeds $150$\,dB effective dynamic range, 
exercising the extreme regime where single-sensor capture saturates. 
DMEB, achieves the best reconstruction quality.}
\label{tab:synthetic_quant}
\vspace{-5mm}
\end{table}

\paragraph{Results on the Synthetic Dataset.}
Figure~\ref{fig:synthetic_result} and table~\ref{tab:synthetic_quant} report qualitative and quantitative results on our synthetic benchmark. 
We compare our method against AFUNet~\cite{li2025afunet}.
As highlighted in the cropped zoom-ins and arrow annotations, AFUNet exhibits residual ghosting near high-contrast boundaries, due to the large exposure gaps between images. 
In contrast, our method produces cleaner reconstructions with sharper structures and fewer halo/bleeding artifacts.
These results indicate that DMEB remains robust even under extreme exposure variations.

\begin{table}[t]
\centering
\scriptsize
\setlength{\tabcolsep}{6pt}
\renewcommand{\arraystretch}{1.08}
\begin{tabular*}{\linewidth}{@{\extracolsep{\fill}}lcc}
\toprule
\textbf{Method} & \textbf{iPhone} & \textbf{Choi~\cite{choi2025dual} 2-view} \\
\midrule
HDR Transformer~\cite{liu2022hdrtransformer} & 23.35 & 26.33 \\
AFUNet~\cite{li2025afunet} & 25.21 & 25.39 \\
\textbf{DMEB (ours)} & \textbf{26.12} & \textbf{33.00}$^\dagger$ \\
\bottomrule
\end{tabular*}
\caption{\textbf{Cross-platform and cross-dataset evaluation.}
We report PSNR-$\mu$ on common-overlap iPhone regions and the external two-view dataset of Choi~\cite{choi2025dual}.
$^\dagger$ denotes zero-shot evaluation without retraining.}
\label{tab:cross_platform}
\end{table}

\paragraph{Results on Consumer and External Datasets.}
We further evaluate DMEB beyond our primary dataset from robotic setup to assess its cross-platform and cross-dataset generality. 
First, we test on the iPhone dataset, 
where a single-shot measurement comprises three images with varying 
fields of view and corresponding LiDAR depth. 
Because the three cameras share only a partial field of view, we evaluate all methods on the common-overlap regions using a valid mask, and compare HDR Transformer, AFUNet, and our three-view DMEB (Tab.~\ref{tab:cross_platform}). 
DMEB achieves the highest PSNR, indicating that modern mobile devices 
equipped with standard multi-view cameras and a depth sensor can 
substantially expand their dynamic range using our approach. 
Figure~\ref{fig:process_on_mobile_phone} shows a qualitative iPhone example. 
Second, to verify that DMEB does not require our full custom rig, we apply it in a zero-shot manner to the external two-view depth setting of Choi~\cite{choi2025dual}, without any retraining or fine-tuning. 
DMEB transfers directly to this setting, reaching 33.00\,dB and outperforming HDR Transformer and AFUNet by a clear margin (Tab.~\ref{tab:cross_platform}), confirming that the method generalizes to sensor configurations beyond those used for training.

\begin{figure}[!t]
    \centering
  
    \includegraphics[width=\linewidth]{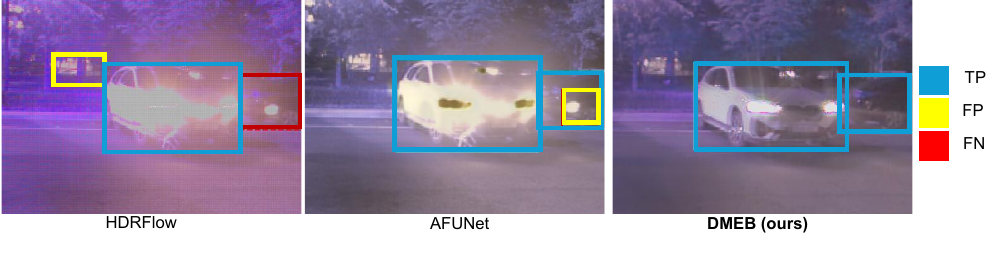}
  \vspace{-6mm}
    \caption{{\textbf{Downstream object detection comparison.}}
    {While strong headlights prevent matching-based methods like HDRFlow and AFUNet from accurately recovering the car's shape, DMEB preserves critical structural details. This improved HDR reconstruction quality from DMEB directly translates to accurate object detection. TP: true positive, FP: false positive, FN: false negative.}}
    \label{fig:obj_det_qual}
\end{figure}

\paragraph{Object Detection with an HDR Image.}
To determine if our high-quality HDR reconstruction translates to improved downstream perception, we evaluate object detection using YOLOv8 on our 90 manually annotated real-world frames. 
DMEB improves Recall/F1 to 0.83/0.73, compared with 0.78/0.63 for HDRFlow and 0.81/0.67 for AFUNet.
Figure~\ref{fig:obj_det_qual} qualitatively demonstrates this advantage: whereas the strong glare from a car's headlights causes HDRFlow and AFUNet to fail at recovering the vehicle's shape, our method successfully preserves the structural details necessary to detect two cars.

\section{Conclusion}
\label{sec:conclusion}
\NEW{
We introduced a multi-view varying-exposure HDR dataset with calibrated depth measurements, collected using a robotic capture platform, an iPhone, and a synthetic renderer.}
\NEW{Together, these data sources cover real robotic scenes, compact consumer-device captures, and controlled extreme-DR synthetic scenes.
}
\NEW{We also presented DMEB as a baseline method, demonstrating that geometry-guided fusion can effectively combine synchronized views captured with large exposure gaps.}
Evaluations on our dataset demonstrate that DMEB validates the effectiveness of this camera rig configuration for robust HDR perception in diverse multi-camera and depth sensor systems.

\paragraph{Future Work.}
A fundamental requirement of our current system is that the multi-view cameras and depth sensors must share overlapping fields of view to effectively expand the dynamic range. To build upon this, future research could explore transitioning from 2D HDR image reconstruction to a comprehensive 3D HDR scene representation, accumulating HDR features directly onto the underlying 3D spatial geometry.

\section*{Acknowledgement}
This work was supported by the National Research Foundation of Korea (NRF) grant funded by the Korea government (MSIT) (RS-2023-00211658); Samsung Electronics Co., Ltd (IO251210-14286-01); and Institute of Information \& Communications Technology Planning \& Evaluation (IITP) grants funded by the Korea government (MSIT) (IITP-2026-RS-2024-00437866 for ITRC, RS-2024-0045788, and RS-2019-II191906 for the Artificial Intelligence Graduate School Program at POSTECH).

\bibliographystyle{splncs04}
\bibliography{references}

\end{document}